\documentclass{tlp}
\usepackage[english]{babel}
\newcommand{\st}{\medskip\noindent}

\def\ni{\noindent}

\newtheorem{definition}{Definition}
\newtheorem{example}{Example}

\def\beq{\begin{equation}}
\def\eeq#1{\label{#1}\end{equation}}
\long\def\COMMENT#1\ENDCOMMENT{\message{(Commented text...)}\par}

\def\and{ \ \wedge}

\def\ar{\leftarrow}

\newcommand{\nin}{\not\in}

\newcommand{\no}{\hbox{\it not}\ }

\newcommand{\ov}{\overline}

\newcommand{\calp}{\mathcal{P}}

\newcommand{\cala}{\mathcal{A}}
\newcommand{\calh}{\mathcal{H}}

\newcommand{\cali}{\mathcal{I}}

\newcommand{\be}{\begin{em}}
\newcommand{\ee}{\end{em}}

\newtheorem{corollary}{Corollary}
\newtheorem{theorem}{Theorem}
\newtheorem{fact}{Fact}

\submitted{11 July 2001}
\revised{11 December 2003. 15 March 2004}
\accepted{30 September 2004}

\begin{document}
\bibliographystyle{acmtrans}

  \title[Normal Forms for Answer Sets Programming]
        {Normal Forms for Answer Sets Programming}

  \author[Costantini and Provetti]
     {STEFANIA COSTANTINI\\
     Dip. d'Informatica, Universit\`a di L'Aquila\\
         Loc. Coppito, L'Aquila, I-67100 Italy\\
         \email{stefcost@di.univaq.it}
         \and ALESSANDRO PROVETTI\\
         Dip. di Fisica, Universit\`a di Messina\\
         Sal. Sperone 31, Messina I-98166 Italy\\
     \email{ale@unime.it}
         }
         
\pagerange{\pageref{firstpage}--\pageref{lastpage}}
\volume{\textbf{X} (Y):}
\jdate{today}
\setcounter{page}{1}
\pubyear{2004}

\label{firstpage}

\maketitle

%%%%%%%%%%%%%%%%%%%%%%%%%%%%%%%%%%%%%%
\begin{abstract}
Normal forms for logic programs under stable/answer set semantics
are introduced. We argue that these forms can simplify the study
of program properties, mainly consistency.
The first normal form, called the {\em kernel} of the program, is useful
for studying existence and number of answer sets.
A kernel program is composed of the atoms which are undefined
in the Well-founded semantics, which are those that directly affect
the existence of answer sets.
The body of rules is composed of negative literals only.
Thus, the kernel form tends to be significantly more
compact than other formulations.
Also, it is possible to check consistency
of kernel programs in terms of colorings of the Extended Dependency Graph
program representation which we previously developed.
The second normal form is called {\em 3-kernel.} A 3-kernel program
is composed of the atoms which are undefined
in the Well-founded semantics. Rules in 3-kernel programs have at most
two conditions, and each rule either belongs to a cycle, or defines a connection
between cycles. 3-kernel programs may have positive conditions.
The 3-kernel normal form is very useful for the static analysis of program consistency, i.e., the syntactic characterization of existence of answer sets.
This result can be obtained thanks to a novel graph-like representation of programs, 
called Cycle Graph which presented in the companion article \cite{Cos04b}.
\end{abstract}


%%%%%%%%%%%%%%%%%%%%%%%%%%%%%%%%%%%%%%%%%%%%%%%%%%%%%%%%%%%%%
\section{Introduction}\label{motivations}

Answer Set Programming is based on the Answer Set
semantics \cite{GelLif88} \cite{GelLif91}, which  gives a declarative
meaning to negation as failure and establishes a direct
connection to Reiter's Default logic and other relevant
non-monotonic reasoning formalism.
In Answer Set Programming (from now on, ASP), solutions to a problem are
represented by answer sets, and not by variable substitutions
produced in response to a query, like in traditional logic
programming.
The Answer Set semantics deals
correctly with cyclic negative definitions, by selecting some of
the classical models of the theory (in particular, those that are
minimal and supported).
In contrast, in the Well-founded Semantics
\cite{wfm} all atoms involved in negative cycles are deemed {\em
undefined}.

Our long-term research objectives, to which this article
contributes the foundations, are:

\begin{enumerate}
\item to study the property of consistency (existence of an answer set) of a
  program;
\item to develop some practical model theory for ASP, e.g., finding the
  syntactic features that affect existence and number of answer sets;
\item to investigate the relationship between the restricted syntax of normal forms and the development of fast algorithms for ASP computation.
\end{enumerate}

In this article we propose two normal forms for logic
programs. Both normal forms have a
uniform restricted syntax, i.e., no facts and few positive conditions.
Hence the representation of programs is rather simple.

First, we define the \be kernel \ee normal form, which forbids positive
conditions, facts (and their direct consequences) and undefined atoms, i.e.,
atoms that can be considered irrelevant from the point of view of program consistency.

The study of program consistency, as well as actual model
computation can be made much easier by the \be 3-kernel \ee
normal form, where the length of rule bodies is limited to two
literals, and some further simplification of irrelevant atoms and
rules is made (at the expense of reintroducing
some positive conditions). We define a simple transformation from
kernel to 3-kernel programs.

The results presented here build on our earlier
results on program representation and analysis under the answer set
semantics.
First, \cite{BCDP99} proposed an approach where
programs are represented by directed graphs, and
deduction algorithms are given in terms of graph coloring.
The 3-kernel normal form will make the corresponding graphs have a regular and simple structure.
Such feature greatly helps
when checking consistency, as shown in \cite{Cos04b}, which gives
the first purely-syntactic and complete characterization of
consistent (w.r.t. Answer Sets semantics) logic programs.

%%%%%%%%%%%%%%%%%%%%%%%%%%%%%%%%%%%%%%%%%
% Left out for space reasons
\COMMENT
%%%%%%%%%%%%%%%%%%%%%%%%%%%%%%%%%%%%%%%%%%%%%%%%%%%%%%%%%%%%%
\subsubsection*{Normalization}\label{motiv-normalization}
The success of the Binary Decision Diagram (BDD)
representation of propositional formulae, which is adopted in a
number of implementation efforts, e.g., in planning,
provides motivations to pursue the introduction of normal form in ASP.
Indeed, \textit{Ordered BDD} is a canonical
form, i.e., each formula has only one representation.
Whenever a canonical form is not easy to define, quasi-canonical, or \be
normal \ee forms can also be useful. The difference with a proper
canonical form is that every expression has a normal form, but
there may be different expressions in normal form that are
equivalent.
A canonical form may not be a suitable idea for ASP as several logic programs
may have the same answer sets.
However, w
\ENDCOMMENT

We believe that several, alternative normal forms can be introduced and studied to
improve the body of technical results and useful transformations for ASP programs.
The contributions of this article are, in summary:

\begin{enumerate}
\item the definition of the kernel format;

\item a representation theorem for the kernel form;

\item the definition of 3-kernel normal form;

\item a normalization algorithm for 3-kernel form.
\end{enumerate}

%%%%%%%%%%%%%%%%%%%%%%%%%%%%%%%%%%%%%%%%%%%%%%%%%%%%%%%%%%%%%
\section{The kernel Normal Form}\label{kernelform}

We assume the standard definitions of (propositional) general logic
program (henceforth, {\em program}) and of Well-founded
\cite{wfm}, stable model \cite{GelLif88} and answer set semantics
\cite{GelLif91}.
Whenever we mention consistency (or stability)
conditions, or the Gelfond-Lifschitz transformation, we refer to
those introduced in \cite{GelLif88}.

Let $\Pi$ be a logic program, we denote by $WFS(\Pi)$ the
well-founded model of $\Pi$. $\Gamma(\Pi,S)$ denotes the
application of the Gelfond-Lifschitz operator to $\Pi$ w.r.t. the
set of literals $S$.

%%%%%%%%%%%%%%%%%%%%%%%%%%%%%%%%%
\begin{definition}
A program $\Pi$ is WFS-irreducible if  and only if
$WFS(\Pi)=\langle\emptyset,\emptyset\rangle$.
\end{definition}

\ni
That is, in WFS-irreducible programs all the atoms have
truth value {\em undefined} under the Well-founded semantics.

The atoms that are relevant for deciding whether answer sets exist
and finding them \cite{Cos95} are exactly those that are deemed
undefined under the Well-founded semantics. Therefore, we aim for
normal forms that are WFS-irreducible.

\ni
Below is the definition of kernel normal form.

%%%%%%%%%%%%%%%%%%%%%%%%%%%%
\begin{definition}
A logic program $\Pi$ is in kernel normal form (or, equivalently,
$\Pi$ is a kernel program) if and only if the following conditions
hold.

\begin{enumerate}
\item $\Pi$ is WFS-irreducible;
\item every rule has its body composed of negative literals only;
\item every atom occurring in $\Pi$ appears in the body of some rule;
\end{enumerate}
\end{definition}

\ni
It is easy to see that: (a) in kernel programs there are no facts and (b) every atom occurs
as the head of some rule and also, by definition, in the body of some
rule (possibly the same one). Moreover,
one can observe that all atoms are either part of a cyclic
definition or are defined by using atoms that are part of a cycle. In
other words, all atoms {\sl somewhat} depend on cycles.
This notion is made precise and developed in the work of \cite{Cos04b}.

%%%%%%%%%%%%%%%%%%%%%%%%%%%%%%
\begin{example}\label{ex:3col}
Consider the problem of 3-colorability of a graph.
Given a graph ${\cal G}$, we need to show a complete assignment
of nodes to one of three colors (in this case green, red and blue)
such that no adjacent nodes are assigned to the same color.
The program below solves the 3-colorability problem and is in
kernel normal form.

Let us name nodes with integers.
For each node (here we consider node $0$) the program must contain these 3 rules, which impose the assignment of a color to the considered node:

\st $
\begin{array}{l}
color(0,\ red)   \ar \no color(0,\ blue), \no color(0,\ green).\\
color(0,\ blue)  \ar \no color(0,\ red),  \no color(0,\ green).\\
color(0,\ green) \ar \no color(0,\ blue), \no color(0,\ red).
\end{array} $

\st
The next set of rules is used to \textit{record} the color
assignments that have not been chosen:

\st $
\begin{array}{l}
n\_color(0,\ red)   \ar \no color(0,\ red).\\
n\_color(0,\ green) \ar \no color(0,\ green).\\
n\_color(0,\ blue)  \ar \no color(0,\ blue).
\end{array} $

\st
Finally, for each edge of ${\cal G}$ we need to add to the program
the following set of rules (the sample set below is for an edge between vertex $0$
and vertex $1$):

\st $
\begin{array}{l}
edge\_ok(0,\ 1) \ar \no edge\_ok(0,\ 1).\\
edge\_ok(0,\ 1) \ar \no edge\_ko(0,\ 1).\\
\\
edge\_ko(0,\ 1) \ar \no n\_color(0,\ red),   \no n\_color(1,\ red).\\
edge\_ko(0,\ 1) \ar \no n\_color(0,\ green), \no n\_color(1,\ green).\\
edge\_ko(0,\ 1) \ar \no n\_color(0,\ blue),  \no n\_color(1,\ blue).
\end{array} $

\st
The first two rules are used to impose the truth of $edge\_ok(0,\ 1)$
and, equivalently, the falsity of $edge\_ko(0,\ 1)$%
\footnote{Normally, ASP solvers extend the language of logic programs with
abbreviations that allow to specify concisely a constraint on the value
of a formula. So, the two rules above are often abbreviated by the formula $\ar edge\_ko(0,\ 1)$.}.

It remains easy to show that each and all answer sets of the program defined
as in the schema above contain a set of \textit{color} atoms, one for each node, from which a solution
to the original problem, i.e., a suitable 3-coloration of the graph,
can be read out.
Similarly, we can show that if the considered graph admits a 3-coloring then
the correspondent kernel program defined using the schema above must admit a
corresponding answer set.
\end{example}
%%%%%%%%%%%%%

\ni
We are now ready to state and prove the main technical result, i.e,
that kernel is a normal form.

%%%%%%%%%%%%%%%%%%%%%%%%%%%%%%%%%%%%%%%%%%%%%%%%%%%%%%%%%%%%%
\section{The proof of normality}\label{proofnormal}
The kernel form is normal in the sense that any logic program
under answer set semantics admits an equivalent kernel program,
i.e., one which has the same answer sets, modulo some projection.

This is a consequence of the following representation Theorem
\ref{rapp-theo}. An alternative, elegant proof has been recently
suggested by W. Marek and J. Remmel in a personal communication.
It consists in showing that their rational reconstruction of
Turing machines \cite{MarRem01} can be kernelized, thus showing
that kernel programs can encode all problems in the NP class.

\ni
In order to state the Theorem, let us recall the following:

\begin{fact}
The answer sets of any logic program form an anti-chain\footnote{A collection
of sets is an anti-chain if no component is subset of another.}.
\end{fact}

\ni
This fact follows directly from minimality of answer sets.
Consistency of the given program is immaterial here. As
\cite{MarTru99} point out, logic programs can be seen as a compact
representation of anti-chains. The representation theorem follows.
For simplicity, it is worded for normal programs, i.e., with no
explicit negation $(\neg)$.

%%%%%%%%%%%%%%%%%%%%%%%%%%%%%%%%
\begin{theorem}\label{rapp-theo}
Let $\calh=\{a_1,\dots a_n\}$ be a set of atoms and let
$\calp(\calh)$ be its powerset. Let $\cala\subseteq \calp(\calh)$
be an arbitrary anti-chain over $\calh$. There exists a kernel
logic program that has exactly the collection of answer sets
$\cala$, modulo projection over $\calh$.
\end{theorem}

\begin{proof}
The proof is by construction.

\ni
First, suppose that $\calh$ contains neither $m$ nor $\bot$;
these are special atoms used in the construction below. Each
component, say $A \in \cala$ is a set $\{a_1,\ldots,a_r\}
\subseteq \calh$. Let $\{n_1,\ldots,n_s\} = \calh \setminus A$ be
the set of atoms not belonging to $A$.

\ni
Second, we \textit{complete} $\calh$ by adding to it, for each atom
$h_i\in \calh$ a {\it fresh} atom $\overline{h_i}$.

\ni
Now, the kernel $\pi_\cala$ is built as follows:

\begin{itemize}
\item [i)]
for each atom $h_i\in \calh$ we put in $\pi_\cala$

\st $
\begin{array}{l}
h_i \ar \no \overline{h_i}.\\
\overline{h_i} \ar \no h_i.
\end{array}  $

\st

\item[ii)]
for each component, $A=\{a_1,\ldots,a_r\}$, of $\cala$ we put in
$\pi_\cala$ the following rule

\st $
\begin{array}{l}
m \ar \no \overline{a_1}, \dots \no \overline{a_r}, \no n_1, \dots
\no n_s.
\end{array}  $

\item[iii)]
Finally, we put the consistency axiom:

\st $
\begin{array}{l}
\bot \ar \no \bot, \no m.
\end{array}  $

\end{itemize}

\st
The intuitive reading is: in order for $\pi_\cala$ to be
consistent, $m$ must be true, so at least one of the rules
defining $m$ must have all its conditions true.
These conditions
describe exactly one of the answer sets. It is easy to see that,
by the anti-chain property which is enforced by the introduction
of the $\overline{a_i}$ atoms, no two rules for $m$ can {\em fire}
under the same answer set.

\ni
In fact, suppose that $\cali$
 is an answer set for $\pi_\cala$ s.t. two rules of $\pi_\cala$ fire, say

\st $
\begin{array}{l}
m \ar \no \overline{a_1}, \dots \no \overline{a_r}, \no n_1, \dots
\no n_s.\\
\\
m \ar \no \overline{a_1'}, \dots \no \overline{a_t'}, \no n_1',
\dots\no n_u'.
\end{array}  $

\st
where we primed the second rule to distinguish its atoms.
Now, by repeated application of Marek and Subrahmanian lemma \cite{GelLif91},
we conclude that $a_1,\dots a_r\in\cali$ and $a_1',\dots a_t'\in\cali$.
By the
antichain property, $\{a_1,\dots a_r\}$ and $\{a_1',\dots a_t'\}$
are disjoint.
Thus, for each $a_i$ we have $a_i\nin\{a_1',\dots a_t'\}$ and then by
construction $\overline{a_i}\in\cali$; but this means that
condition $\no n_x$ with $n_x\equiv a_i$ is false. Therefore, the
second rule cannot fire.
A similar argument applies for an
arbitrary $a_j'\nin\{a_1,\dots a_r\}$.

So, each answer set corresponds to the truth of exactly one body
of the definition of $m$.
In turn, each such body corresponds, up to $\calh$ (i.e., in all but the
special atom \textit{m}), to a component of $A$, as intended.

\ni
It remains easy to check that for any $\cala$, the resulting
$\pi_\cala$ is in kernel format.
This concludes the proof.
\end{proof}

%%%%%%%%%%%%%%%
\begin{corollary}
Any logic program $\Pi$ admits an equivalent ---w.r.t. answer set semantics---
program in kernel format.
\end{corollary}

The theorem above establish that kernel program is a normal form,
but it does not suggest a straightforward, efficient computational
mechanism for answer set computing.
The main reason is that what
is taken into account, i.e., anti-chains over $\calh$, may be of
cardinality exponential w.r.t. that of $\calh$.
However, it is important as far as establishing that kernel
 equivalents always do exist, and hence that Kernel is a normal form.

%%%%%%%%%%%%%%%%%%%%%%%%%%%%%%%%%%%%%%%%%%%%%%%%%%%%%%%%%%%%
\COMMENT
%%%%%%%%%%%%%%%%%%%%%%%%%%%%%%%%%%%%%%%%%%%%%%%%%%%%%%%%%%%%
\subsection{From normal to kernel programs}\label{kernelization}
We are interested in using the kernel normal form as a standard presentation
of programs to solvers.
That is, to take arbitrary programs and
transform them into kernel equivalents before calling the ASP solver on them.
This activity is called kernelization.
This is what happens in the propositional satisfiability scenario where the conjunctive
normal form (CNF) is the standard presentation of formulae to the satisfiability checkers.

The idea behind kernelization is that solvers could exploit the restricted syntax of kernels to compute Answer Sets efficiently.
\cite{Mag02} has developed a version of the  \textit{smodels} solver that takes as input kernel programs; therefore, there is no need to
handle positive conditions, undefined atoms etc.%
\footnote{A preliminary benchmark of the modified smodels suggests that --on kernel programs-- it
can run around 50\% faster than the standard version, even though no specific heuristics has been added yet.}

Kernelization, i.e., transformation into kernel, is of practical interest only if it can be done efficiently, i.e., in low polynomial time.
Indeed, complexity theory suggests that there should be a translation from normal programs to kernel programs such the latter program can only be \textit{polynomially larger} than the original one.

We are currently developing a kernelization algorithm based on unfolding, but here we
content ourselves with giving a \textit{quadratic} upper bound on the expansion of the program size that kernelization can give.

Let $n$ and $m$ be respectively the number of atoms and rules appearing in a given program $\Pi$.
\cite{LinZha03} define a \textit{faithful} translation from normal logic programs to propositional formulae, say $\Phi$.
that introduces at most $k=n^2+m$ atoms and $l=mn$ rules.
Notably, to achieve faithfulness their translation conditions the program instance to rid
it from positive cyclic definitions.
Now, consider the following translation from SAT formulae to kernel logic programs, say $\Pi_k$:

\begin{itemize}
    \item for each atom (propositional variable) $a$ in $\Phi$ put in $\Pi_k$ two rules:

    $a \ar \no \ov{a}$ and $\ov{a} \ar \no a$.

    \item for each rule, say $h\ar p, \no n$ put in $\Pi_k$ the constraint: $\ar \no h, \no \ov{p}, \no n$
\end{itemize}

\ni
This back-translation creates $\Pi_k$ with $2k=2(n^2+m)$ atoms and $l=(mn)$ rules.
It is easy to show that the back-translation to kernel is faithful and so is the composition
of the Lin-Zhao's translation to our.
As a result, we find a faithful translation from normal logic programs to kernels that
adds at most a quadratic number of atoms and a linear number of rules.
%%%%%%%%%%%%%%%%%%%%%%%%%%%%%%%%%%%%%%%%%%%%%%%%%%%%%%%%%%%%
\ENDCOMMENT
%%%%%%%%%%%%%%%%%%%%%%%%%%%%%%%%%%%%%%%%%%%%%%%%%%%%%%%%%%%%

%%%%%%%%%%%%%%%%%%%%%%%%%%%%%%%%%%%%%%%%%%%%%%%%%%%%%%%%%%%%%
%%%%%%%%%%%%%%%%%%%%%%%%%%%%%%%%%%%%%%%%%%%%%%%%%%%%%%%%%%%%%
%%%%%%%%%%%%%%%%%%%%%%%%%%%%%%%%%%%%%%%%%%%%%%%%%%%%%%%%%%%%
\section{The 3-kernel normal form}\label{3-kernel}
In this Section we define a variation
of the kernel normal form that we call 3-kernel normal form, where
rules have at most two conditions.
The more restrictive 3-kernel form has in our view two advantages, i.e., it

\begin{enumerate}
    \item  allows the study of program consistency in a mathematically elegant way and

    \item might help to discover efficient algorithms and heuristics for answer set computation.
\end{enumerate}

Also, we hope that it will become a useful program representation in the search for new
algorithms/heuristics. We expect the 3-kernel normal form to play for ASP the same
r\^ole that 3SAT
plays for propositional logic, i.e., a streamlined formula presentation that
is generally used for the input of satisfiability solvers and model checkers.
In our related work \cite{BCDP99} \cite{Cos04b}, we have argued that the
existence of answer sets depends on both the cycles which are
present in the program, and their connections.
Indeed, the 3-kernel form is defined with the aim of making cyclic definitions and
their connections thereof explicit.
Before defining the 3-kernel normal form, it is
useful to report some definitions from \cite{Cos04a}.

%%%%%%%%%%%%%%%%%%%%%%%%
\begin{definition}
\label{cycle}
A set of rules $C$ is called a cycle if it has the following form:

\st $
\begin{array}{l}
\lambda_1 \ar \no \lambda_2, \Delta_1\\
\lambda_2 \ar \no \lambda_3, \Delta_2\\
\dots\\
\lambda_n \ar \no \lambda_1, \Delta_n
\end{array} $

\st
where $\lambda_1,\ldots ,\lambda_n$ are distinct atoms. Each
$\Delta_i$, $i \leq n$, is a possibly empty conjunction
$\delta_{i_1}, \ldots, \delta_{i_h}$ of conditions (positive or
negative), where $i_h \geq 0$
%%%%%%%%%%%%%%%%%%%%%%%%%%%%%%%%%
% {\sc Ale: why are empty deltas disallowed?}
%%%%%%%%%%%%%%%%%%%%%%%%%%%%%%%%%
and for each literal $\delta_{i_j}\in \Delta_i$, $i_j
\leq i_h$, $\delta_{i_j} \neq \lambda_i$ and $\delta_{i_j} \neq
\no \lambda_i$.
The $\Delta_i$'s are called {\em AND handles} of the cycle.
\end{definition}

We say that $\Delta_i$ from Definition above is an {\em AND
handle} for atom $\lambda_i$, or, equivalently, an AND handle
referring to $\lambda_i$.
Cycles of length $n=1$ correspond to
self-loops $\lambda_1 \ar \no\lambda_1,\Delta_1$, which are
critical to determine consistency.
We will say that
$C$ has size $n$ and it is even (respectively odd) if $n=2k$, $k
\geq 1$ (respectively, $n=2k+1$, $k \geq 0$).
The $\lambda$'s are
the atoms \be involved \ee in cycle $C$, or, equivalently, the \be
composing atoms \ee of the cycle.
Rules belonging to a cycle $C$
are said to be \be involved in, \ee or \be belong to, \ee or \be
form \ee the cycle $C$.

\begin{definition}\label{or-one-handle}
A rule is called an \be auxiliary rule of cycle \ee $C$
(or, equivalently, \be to \ee cycle $C$)
if it is of this form:

\ni
$\lambda_i \ar \Delta$

\ni
where $\lambda_i$ is involved in cycle $C$, and $\Delta$ is a
non-empty conjunction $\delta_{i_1}, \ldots, \delta_{i_h}$ of
literals (either positive or negative), where $i_h \geq 1$ and for
each $\delta_{i_j}$, $i_j \leq i_h$, $\delta_{i_j} \neq \lambda_i$
and $\delta_{i_j} \neq \no \lambda_i$. $\Delta$ is called an OR
handle of cycle $C$ (more specifically, an OR handle for
$\lambda_i$ or, equivalently, and OR handle referring to
$\lambda_i$).
\end{definition}

A cycle may have several auxiliary rules.
Hence, a cycle may have
some AND handles, occurring in one or more of the rules that form
the cycle itself, and also some OR handles, occurring in its
auxiliary rules.
Handles are seen as connections between cycles.
In order to make these connections explicit, it is useful that:

\begin{itemize}
\item each handle be composed of just one atom, not belonging to the cycle, and
\item bridges should be as short as possible,
even at the expense of re-introducing some positive literals.
\end{itemize}

\ni
These requirements are fulfilled by the definition of
3-kernel normal form.

\begin{definition}\label{3-kerneldef}
A logic program $\Pi$ is in 3-kernel normal form (or,
equivalently, $\Pi$ is a 3-kernel program) if the following
conditions hold.

\begin{enumerate}
\item $\Pi$ is WFS-irreducible;
\item every atom occurring in $\Pi$ is involved in some cycle;
\item each rule of $\Pi$ is either involved in a cycle,
or is an auxiliary rule of some cycle, or both;
\item the body of each rule of $\Pi$ which is involved
in a cycle consists of either one or two literals;
\item each atom occurring in the AND handle of a cycle
is not involved in that cycle;
\item the body of each rule of $\Pi$ which is an auxiliary rule of some cycle
consists of exactly one literal.
\end{enumerate}
\end{definition}

In fact, as mentioned above
all atoms in a kernel program are either part of a cyclic
definition or defined (directly or indirectly) using atoms that are part of a cycle.
However, allowing handles composed of several atoms means that
a handle constitutes a connection between the cycle it refers to,
and several other cycles.
Also, connections between cycles can be defined indirectly, by means of
a chain of dependencies.

In the view of studying program consistency
in terms of a \textit{Cycle Graph} \cite{Cos04b}, where vertexes are cycles and
edges are connections between cycles,
the 3-kernel form guarantees that each handle
consists of just one condition, and thus it corresponds {\em
exactly} to a connection between cycle $C$ and  one other cycle
$C'$. This leads to a much cleaner representation of the program.

%%%%%%%%%%%%%%%%%%%%%%%%%%%%%%%%%%%%%%%%%%%%%%%%%%%%%%%%%%%%%%%%%%%%%
\subsection{3-kernelization}
The aim of this section is to show how to transform a kernel program
into a 3-kernel program.
%That is, in previous section we have
%transformed a logic program $\Pi$ into $\Pi' = ker(\Pi)$.
%Now, we want to transform $\Pi'$ into $3ker(\Pi)$.
Given a kernel program, every rule with non-unit body can
be eliminated, by transforming it into a cycle.
Consider for instance program $\pi_5$:

\st $
\begin{array}{ll}
p \ar \no p.\\
p \ar \no a, \no c.\\
a \ar \no b.\\
b \ar \no a.\\
c \ar \no d.\\
d \ar \no c.
\end{array} $

\st
Program $\pi_5$ would be in 3-kernel format but for the second
rule. In fact, it is an auxiliary rule of cycle
 $\{p \ar \no p\}$, but it does not 
match the definition of 3-kernel as its body consists of two literals instead
of one. This rule is then transformed as follows:

\st $
\begin{array}{ll}
p \ar \no p.\\
p \ar \no h_1.\\
h_1 \ar \no h_2.\\
h_2 \ar \no h_3, \no a.\\
h_3 \ar \no h_4.\\
h_4 \ar \no h_5, \no c.\\
h5 \ar \no p.\\
a \ar \no b.\\
b \ar \no a.\\
c \ar \no d.\\
d \ar \no c.
\end{array} $

\st
The latter program is equivalent to $\pi_5$ up to the language
of $\pi_5$ itself. The long 
auxiliary rule has been replaced by a new cycle.
More generally, the following transformation
can be applied.

%%%%%%%%%%%%%%%%%%%%%%%%%%%%%%%%%%%%%%%%%%%%%
\begin{definition}[Long Rules simplification]
Let $\Pi$ be a program in kernel normal form.
The Long Rule simplification $\Pi'$ of $\Pi$ is created as follows.
Each rule

\ni
$\rho:\ h \ar \no b_1,\ldots,\no b_j$

\ni
occurring in $\Pi$ that is either

\begin{enumerate}
\item
auxiliary to a cycle $C$, with $j > 1$, or

\item involved in a cycle $C$, and $j > 2$
\end{enumerate}

\ni
is substituted in $\Pi'$ by the set of rules (that form a new cycle):

\st $
\begin{array}{ll}
h \ar \no h_1.\\
h_1 \ar \no h_2.\\
h_2 \ar \no h_3, \no b_1.\\
h_3 \ar \no h_4.\\
h_4 \ar \no h_3, \no b_2.\\
\ldots \\
h_{2j} \ar \no h_{2j+1}, \no b_j.\\
h_{2j+1} \ar \no h.
\end{array} $

\ni
where $h_1,\ldots,h_{2j+1}$ are fresh atoms, i.e., do not appear in $\Pi$.
\end{definition}

By discarding the-truth value of the fresh atoms, one can check that the
truth conditions for $h$ remain the same as before.

%The result is a program with additional cycles.
As far as complexity
is concerned, we notice that a long rule has at most $n=|\cala|$
literals in the body, where $|\cala|$ is the number of atoms
occurring originally in $\Pi$.
For each condition
appearing in the original rule we introduce two new atoms
and two new rules.
Then we introduce a final rule to \textit{close} the new cycle.
Thus, for each long rule we
add at most $2n$ atoms and $2n + 1$ rules.
In the worst case, i.e., when we have to apply the transformation
to all rules of $\Pi$, we add $n \cdot (2 \cdot n)$ new atoms and
$n \cdot ((2 \cdot n) + 1)$ new rules.

\ni
Since every atom occurring
in $\Pi$ has at least one defining rule, then $\cala$ is less then
or equal to $m=|\Pi|$, i.e., to the number of rules of $\Pi$.
Hence, the
new program after Long rule simplification has at most
$2m^2$ new atoms and
$2m^2 + m$ new rules.

%%%%%%%%%%%%%%%%%%%%%%%%%%%%%%%%%%%%%%%%%%%%%%%%%%%%%%%%%%%%
\subsection{Bridge elimination and other useful equivalences}
Even after performing the previous transformation, the program may
still not be in 3-kernel form. In fact, it may contain rules that
do not belong to any cycle nor are they auxiliary to a cycle. Such
rules are said to from \be bridges, \ee seen as {\em paths}
connecting cycles. Bridges can be eliminated without affecting the
semantics, at the price of dropping some atoms. The truth values
of the dropped atoms can be reconstructed at a later stage since
 it can be proved that the truth value of any single atom of a bridge
 determines the truth values of the each other atom of the bridge.
Elimination of bridges will now be discussed by case analysis.
In cases (i) and (ii) we discuss how to eliminate bridges
that originate in OR handles (called {\em OR-bridges}).
In cases (iii) and (iv) we discuss how to eliminate bridges
that originate in AND handles (call {\em AND-bridges}).

\ni
(i) Consider a set of rules of the form:

\st $
\begin{array}{ll}
p \ar \no p.\\
p \ar \no e.\\
e \ar \no f.\\
f \ar \no a.\\
a \ar \no b.\\
b \ar \no a.
\end{array} $

\ni
that corresponds to a bridge between cycles $\{p \ar \no p\}$
and $\{a \ar \no b.\ b \ar \no a.\}$ via an OR handle. In fact,
$p$ depends on $\no e$ (first rule, or first step, of the bridge),
$e$ depends on $\no f$ (second step), $f$ depends on $\no a$
(third step). 
Since the bridge originates in an OR handle, it will be called an {\em OR-bridge}.
The bridge involves three rules, i.e. three dependencies.
Apart from $p$,
but it involves two atoms, namely $e$ and $f$.
Based on the number of atoms, we say that it is an OR-bridge of \be even length. \ee 
To the extent of
checking consistency, this set of rules can be substituted by the
following set.

\st $
\begin{array}{ll}
p \ar \no p.\\
p \ar \no a.\\
a \ar \no b.\\
b \ar \no a.
\end{array} $

\ni
Clearly, the latter program is equivalent to the former up to
the language. 
In the latter one however there is no bridge, 
rather a direct connection between the
two cycles via the OR handle {\em \no a}.
The truth value of atoms $e$ and $f$ can be easily
reconstructed from each answer set of the latter program.
Moreover, the transformation seen above can be applied whenever an
even number of atoms are used to \textit{form} a bridge.

\ni
(ii) Consider a set of rules of the form:

\st $
\begin{array}{ll}
p \ar \no p.\\
p \ar \no e.\\
e \ar \no f.\\
f \ar \no g.\\
g \ar \no a.\\
a \ar \no b.\\
b \ar \no a.
\end{array} $

\ni
where we can see an OR-bridge of odd-length 
between cycles $\{p \ar \no p\}$ and $\{a \ar \no b.\ b
\ar \no a. \}$ involving atoms $e, f,$ and $g$.
To the extent of checking consistency, this set can be substituted by the
following:

\st $
\begin{array}{ll}
p \ar \no p.\\
p \ar a.\\
a \ar \no b.\\
b \ar \no a.
\end{array} $

\ni
The two set of rules, seen as programs, are equivalent. The
price we pay to ridding programs of this type of rules is the
introduction of a positive condition.
The formal definition of OR-bridge follows.

\begin{definition}[OR-bridge]
An OR-bridge is a set of rules

\st $
\begin{array}{ll}
p \ar \no q.\\
p \ar \no \lambda_1.\\
\lambda_1 \ar \no \lambda_2.\\
\ldots \\
\lambda_n \ar \no a.
\end{array} $

where the first rule belongs to some cycle $C$ where $p$ and $q$
are defined, the second rule is auxiliary to $C$ and $a$ is
defined in some other cycle. An OR-bridge is of \be even (resp.
odd) length, \ee (even OR-bridge for short) if $n$ is even (resp.
odd).
\end{definition}

\begin{definition}[even OR-bridge simplification]
An even OR-bridge can be substituted by the set of rules:

\st $
\begin{array}{ll}
p \ar \no q.\\
p \ar \no a.\\
\end{array} $

\end{definition}

\begin{definition}[odd OR-bridge simplification]
An odd OR-bridge can be substituted by the set of rules:

\st $
p \ar \no q.\\
p \ar a.
$

\end{definition}

\ni
(iii) Consider a set of rules of the form:

\st $
\begin{array}{l}
p \ar \no p,\no e\\
e \ar \no f.\\
f \ar \no a.\\
a \ar \no b.\\
b \ar \no a.
\end{array} $

that corresponds to a bridge between cycles $\{p \ar \no p\}$
and $\{a \ar \no b.\ b \ar \no a.\}$ via an AND handle. In fact,
$p$ depends on $\no e$ (first rule, or first step, of the bridge),
$e$ depends on $\no f$ (second step), $f$ depends on $\no a$
(third step). 
Since the bridge originates in an AND handle, it will be called an {\em OR-bridge}.
The bridge involves three rules, i.e. three dependencies.
Apart from $p$, it involves two atoms, namely $e$ and $f$.
Based on the number of atoms, we say it is an AND-bridge of \be even length. \ee 
In an AND-bridge, the first dependency that forms
the bridge occurs in a rule belonging to a cycle,
rather than in an auxiliary rule like for an OR-bridge.
To the extent of
checking consistency, this set can be substituted by the following
set.

\st $
\begin{array}{ll}
p \ar \no p,\no a.\\
a \ar \no b.\\
b \ar \no a.
\end{array} $

\ni
Clearly, the transformation is equivalence-preserving up to
the language, while the truth value of omitted atoms (namely the
intermediate atoms of the bridge: $e$ and $f$) can be
deterministically obtained from the previous ones.
In the latter program however there is no bridge, 
rather a direct connection between the
two cycles via the AND handle {\em \no a}.

\ni
(iv) Consider a set of rules of the form:

\st $
\begin{array}{ll}
p \ar \no p,\no e\\
e \ar \no f.\\
f \ar \no g.\\
g \ar \no a.\\
a \ar \no b.\\
b \ar \no a.
\end{array} $

where we can see an AND-bridge of odd length between cycles $\{p \ar
\no p\}$ and $\{a \ar \no b. b \ar \no a. \}$, involving atoms $e, f, g$. 
To the extent of
checking consistency, this set of rules can be substituted by the
following set.

\st $
\begin{array}{ll}
p \ar \no p, a.\\
a \ar \no b.\\
b \ar \no a.
\end{array} $

\st Let us make intuition about AND bridges formal.

\begin{definition}[AND-bridge]
An AND-bridge is a set of rules

\st $
\begin{array}{l}
p \ar \no q, \no \lambda_1.\\
\lambda_1 \ar \no \lambda_2.\\
\ldots \\
\lambda_n \ar \no a.
\end{array} $

where the first rule belongs to cycle a $C$, the second rule is
auxiliary to $C$ and atom $a$ is defined in another cycle $C_2$.
An AND-bridge is of \be even (resp. odd) length \ee (even
AND-bridge, for short) if $n$ is even (resp. odd).
\end{definition}

\begin{definition}[even AND-bridge simplification]
An even AND-bridge can be substituted by the set of rules:

\st $
\begin{array}{l}
p \ar \no q, \no a.
\end{array} $

\end{definition}

\begin{definition}[odd AND-bridge simplification]
An odd AND-bridge can be substituted by the set of rules:

\st $
\begin{array}{l}
p \ar \no q, a.
\end{array} $

\end{definition}

The OR and AND bridge simplification can clearly be done in
reasonable time, and have the effect of decreasing the size of the
program.
After applying the above transformations, we have
obtained the 3-kernel normal form of the original program $\Pi$.
From the above reasoning, we are lead to the following general conclusions.

%%%%%%%%%%%%%%%%%%%%%%%%%%
\begin{theorem}
For a given program $\Pi$, its 3-kernel normal form $\Pi =
3ker(\Pi)$ is obtained from its kernel normal form $ker(\Pi)$ via
the application of (i) long rule simplification and (ii) bridge
simplification.
\end{theorem}

%%%%%%%%%%%%%%%%%%%%%%%%%%
\begin{theorem}
For any given program $\Pi$, the answer sets of the 3-kernel
normal form $3ker(\Pi)$ correspond, up to the language, to
those of $\Pi$.
\end{theorem}

Therefore, the answer sets of an arbitrary program $\Pi$ can be obtained
by applying 3-kernelization and then {\em expanding} each answer set over the
language of $\Pi$.
Notice that the bridge simplification reintroduces positive atoms.
The 3-kernel form in fact admits positive handles.
The different kinds of bridges between cycles that
a program may contain after the simplification
introduced above, are illustrated by the following example.

%%%%%%%%%%%%%%%%%%%%%%%%%%%%%%
\begin{example}\label{bridges}
Let $\pi_6$ be the 3-kernel program:

\st $
\begin{array}{ll}
a \ar \no b.\\
b \ar \no a.\\
p \ar \no p,\no b.\\
q \ar \no q.\\
q \ar \no a.
\end{array} $

We can distinguish the even cycle $EC_1 = \{ a \ar \no b.\ b \ar \no a \}$
and the odd cycles $OC_1 = \{ p \ar \no p \}$ and $OC_2 = \{ q \ar
\no q \}$.
There is an AND-bridge from $EC_1$ and $OC_1$, corresponding to the AND handle
$\no b$ of $OC_1$.
There is an OR-bridge from $EC_1$ and $OC_2$,
corresponding to the OR handle $\no a$ of $OC_2$.
In these handles, that form  \textit{bridges,} is the key of the
existence of the answer set $\{b, q\}$.
In fact, the odd cycles if taken alone would be inconsistent.
The even cycle instead has the two answer sets $\{a \}$, $\{b \}$.
By choosing $\{b \}$, the odd cycles are \be constrained \ee by the handles
and become consistent.
%The \textit{Cycle Graph} of this program \cite{Cos04b} would \st
\end{example}

%%%%%%%%%%%%%%%%%%%%%%%%%%%%%%%%%%%%%%%%%%%%%%%%%%%%%%%%%%%%%
\section{Discussion}\label{cyclegraph}
The study of consistency of ASP programs led us to the definition
of kernel and 3-kernel normal forms. Together with the
introduction of the Extended Dependency Graph (EDG) program
representation in \cite{BCDP99}, these normal forms, and the
formal results presented here, provide a suitable theoretical
framework for the study of program properties and the development
of an ASP model theory.

It could be noticed that stating 3-colorability, an NP-complete
problem, as a kernel program (whose size is polynomial in the size of the considered graph)
already amounts to a proof of NP completeness for kernel.
However, such
equivalence is indirect and not suggestive of the connection existing between a general programs and its kernel counterparts.

Indeed, the kernel normal form may yield advantages in terms of
design and implementation of ASP computations.
First, regularities
are exploited in a more concise program representation.
This fact has been exploited, for instance, in the application of
 neuro-genetic approximation methods to ASP
computation in \cite{BGPKT01}.
Second, existing polynomial-time
program simplifications are efficiently integrated in the answer
set computation process and this process is led by semantics
considerations throughout.
Finally, the 3-kernel form, is reminiscent
of {\em predicate binarization,} a parallelization technique that has been
in studied in the context of parallel Prolog program execution, and
 suggests an approach to the parallel computation of answer sets, an
 issue that only recently has received attention.

A subject of future research is to investigate whether relevant
open problems, such as program equivalence, and {\sl safe} program
composition, can be usefully rephrased in terms of 3-kernel
programs.

%%%%%%%%%%%%%%%%%%%%%%%%%%%%%%%%%%%%%%%%%%%%%%%%%%%%%%%%%%%%%

%%%%%%%%%%%%%%%%%%%%%%%%%%%%%%%%%%%%%%
%%%%%%%%%%%%%%%%%%%%%%%%%%%%%%%%%%%%%%
%%%%%%%%%%%%%%%%%%%%%%%%%%%%%%%%%%%%%%
\label{lastpage}
\end{document}